\documentclass[11pt|a4paper]{article}
\usepackage{acl2015}
\usepackage{times}
\usepackage{url}
\usepackage{latexsym}
\usepackage{booktabs}
\usepackage{mdwlist}

\title{USFD: Twitter NER with Drift Compensation and Linked Data}

\author{Leon Derczynski\\
  University of Sheffield, UK\\
  {\small {\tt leon@dcs.shef.ac.uk} } \And
  Isabelle Augenstein\\
  University of Sheffield, UK\\
  {\small {\tt isabelle@dcs.shef.ac.uk} } \And
  Kalina Bontcheva\\
  University of Sheffield, UK\\
  {\small {\tt  kalina@dcs.shef.ac.uk} } }

\date{}

\begin{document}
\maketitle
\begin{abstract}
This paper describes a pilot NER system for Twitter, comprising the USFD system entry to the W-NUT 2015 NER shared task.
The goal is to correctly label entities in a tweet dataset, using an inventory of ten types.
We employ structured learning, drawing on gazetteers taken from Linked Data, and on unsupervised clustering features, and attempting to compensate for stylistic and topic drift -- a key challenge in social media text.
Our result is competitive; we provide an analysis of the components of our methodology, and an examination of the target dataset in the context of this task.
\end{abstract}

\section{Introduction}

Social media is a very challenging genre for Natural Language Processing (NLP) \cite{Derczynski2013b}, providing high-volume linguistically idiosyncratic text rich in latent signals,  the correct interpretation of which requires diverse contextual and author-based information. Consequently, this noisy content renders NLP systems trained on more consistent, longer documents, such as newswire, mostly impotent \cite{Derczynski2014b}. 
Suffering from a sustained dearth of annotated Twitter datasets, it may be useful to understand what makes this genre tick, and how our existing techniques and resources can be generalised better to fit such a challenging text source.

This paper has focused on introducing our Named Entity Recognition (NER) entry to the WNUT evaluation challenge~\cite{baldwin2015wnut}, which builds on our earlier experiments with Twitter and news NER \cite{derczynski2014passive,bontcheva2013twitie,Cun02b-short}. In particular, we push data sources and representations, using what is know about Twitter so far to construct a model that informs our choices.
Specifically, we attempt to compensate for entity drift; to harness unsupervised word clustering in a principled fashion; to bring in large-scale gazetteers; to attenuate the impact of terms frequent in this text type; and to pick and choose targeted gazetteers for specific entity types.

\section{Datasets}

The training and development sets provided with the challenge were drawn from the~\newcite{Ritter2011} corpus.
This was a set of 2394 tweets from late 2010, annotated with ten entity types, including the ``other" type.
A later release in the challenge gave a set of 420 tweets from 2015, annotated in the same way ({\em dev\_2015}).
As no other tweet corpora use this 10-class entity model, we stuck with this data for the supervised parts of our approach.

For language modelling, we used a set of 250 million tweets drawn from the Twitter garden hose, which is a fair 10\% sample of all tweets~\cite{kergl}.
These were reduced to just English tweets using langid.py~\cite{lui2012langid}, and then tokenized using the twokenizer tool~\cite{Connor_Krieger_Ahn_2010}, which gives the same tokenization as used in the input and evaluation corpora.

In addition, we used three sources of gazetteers.
The first two were manually created, and covered named temporal expressions~\cite{brucato2013recognising} and first person names~\cite{Cun02b-short}.
The last comprised more recent data, drawn automatically from Freebase as part of a distant supervision approach to entity detection and relation annotation~\cite{augenstein2014relation}.

\section{Method}
The WNUT Twitter NER task required us to address many data sparsity challenges.
Firstly, the datasets involved are simply very small, making it hard to generalise in supervised learning, and meaning that effect sizes cannot be reliably measured.
Secondly, Twitter language is arguably one of the noisiest and idiosyncratic text genres, which manifests as a large number of word types, and very large vocabularies due to lexical variation~\cite{Eisenstein2013}.
Thirdly, the language and especially entities found in tweets change over time, which is commonly referred to as {\em drift}.
The majority of the WNUT training data is from 2010, and only a small amount from 2015, leading to a sparsity in examples of modern language.
Therefore, in our machine learning approach, many of the features we introduce are there to combat sparsity.

\subsection{Unsupervised Clustering}
We use an unsupervised clustering of terms to generate word type features.
The goal of this is to gain a progressive reduction in the profusion of word types intrinsic to the text type.
250 million tweets from 2010-2012 were used to generate 2,000 word classes using Brown clustering~\cite{brown-cl92}.
Typically 1,000 or fewer are used; the larger number of classes was chosen because it helpfully increased the expressivity of the representation~\cite{derczynski2015brown}, while retaining a useful sparsity reduction.
These hierarchical classes were represented using bit depths of 3-10 inclusive, and then 12, 14, 16, 18 and 20, one feature per depth.
The typical levels are 4, 6, 10 and 20, though selection of bit depths to use often yields brittle feature sets~\cite{koo}, and so we leave it to the classifier to decide which ones are useful. 
These choices are examined in our post-exercise investigations into the model, Section~\ref{sec:brown}, and the clusters provided with this paper.
Finally, we also include the Brown class paths for the previous token.

To aid in filtering out common tokens and reducing the impact they may have as e.g. spurious gazetteer matches, we incorporate a term frequency from our language model.
This is applied to terms that are in the top 50,000 found in our garden hose sample, and represented as a feature having a value scaled in proportion to the term's relative frequency, multiplied by 100 to reduce underflows and ensure it has an effective impact.

\subsection{Morpho-Syntactic Features}
To model context, we used reasonably conventional features: the token itself, the uni- and bigrams in a $[-2,2]$ offset window from the current token, and both wordshape (e.g. {\em London} becomes {\em Xxxxxx}) and reduced wordshape ({\em London} to {\em Xx}) features.

We also included a part-of-speech tag for each token.
These were automatically generated by a custom tweet PoS tagger using an extension of the PTB tagset~\cite{Derczynski2013c}.

To capture orthographic information, we take suffix and prefix features of length $[1..3]$.

Capitalisation is notoriously unreliable in tweets, and also often overfitted to by newswire systems trained on more canonical forms of text.
To wean these systems away from capitals while trying to minimise false negatives, we used case-insensitive gazetteers to generate gazetteer features.

\subsection{Gazetteers}
While we collected and experimented with a variety of gazetteers, the most helpful ones were:

\begin{itemize}
\item Freebase gazetteers mined for distant supervision~\cite{augenstein2014relation};
\item ANNIE first name lists~\cite{Cun02b-short};
\item First name trigger terms~\cite{derczynski2014passive};
\item Lists of named temporal expressions~\cite{brucato2013recognising}, used due to the prevalence of festival and event names in the {\em other} category.
\end{itemize}

Freebase~\cite{bollacker2008freebase} is a large knowledge base consisting of around 3 billion facts\footnote{The Freebase project is being discontinued as of May 2015, however, the data is being integrated with Wikidata~\cite{wikidata2014}.~\url{https://plus.google.com/109936836907132434202/posts/3aYFVNf92A1}}.
    As such, it has been used extensively as background knowledge for NLP tasks such as entity and relation extraction~\cite{augenstein2014relation}. Gazetteers for the 10 entity types were retrieved from Freebase semi-automatically. Some of the types correspond to Freebase types directly, e.g. person corresponds to /people/\textit{person}, but for other types such as \textit{product} there are no directly corresponding types. To build gazetteers, we therefore retrieved all Freebase types for all entities in the training corpus and selected the most prominent Freebase types per entity type in the gold standard. The list of Freebase types corresponding to each entity type in the gold standard is listed in Table~\ref{tab:ftypes}.
    
    \setlength{\tabcolsep}{1em}
    \begin{table}[t]
    \fontsize{8}{10}\selectfont
    \begin{center}
    \begin{tabular}{l l}
    \toprule
    \bf NE type & \bf Freebase type \\
    \hline
    company & /business/business\_operation, \\
     & /organization/organization \\
    facility & /architecture/building, /architecture/structure,  \\
     &  /travel/tourist\_attraction \\
    geo-loc &  /location/location \\
    movie & /film/film \\
    musicartist & music/artist \\
    other & /education/university, /time/holiday,  \\
     & /time/recurring\_event \\
    person & /people/person \\
    product &  /business/consumer\_product, /business/brand,\\
     & /computer/software, /computer/operating\_system,\\
     & /computer/programming\_language, \\
     & /digicams/digital\_camera,\\
     & /cvg/computer\_videogame,  /cvg/cvg\_platform,\\
     & /food/food, /food/beverage, /food/tea, /food/beer,\\
     & /food/brewery\_brand\_of\_beer, /food/candy\_bar,\\
     & /food/cheese, /food/dish, /wine/wine \\
     & /distilled\_spirits/distilled\_spirit \\
    
    sportsteam & /sports/sports\_team \\
    tvshow &  /tv/tv\_program \\
    \bottomrule
    \end{tabular}
    \end{center}
    \caption{\label{tab:ftypes} NE types and corresponding Freebase types used for creating gazetteers}
    \end{table}

\normalsize

For each Freebase type, separate gazetteers were created for entity names and alternative names (aliases), since the latter tend to be of lower quality.

There were several other gazetteer sources that we tried but which did not work very well: IMDb dumps,\footnote{See http://www.imdb.com/interfaces} Ritter's LabeledLDA lists~\cite{Ritter2011} (duplicated in the baseline system), and ANNIE's other gazetteers (largely consisting of organisations, locations, and date entities) en masse.
Each of these introduced a drop in performance or an unstable performance, possibly due to the increased ambiguity.
This is a known problem with discriminative learning -- only a certain amount of gazetteers may be used as features in this way before performance of a discriminative learner drops~\cite{smith-osborne:2006:CoNLL-X}.

\subsection{Learning Models and Representation}
As BIO NE chunking is readily framed as a sequence labeling problem, we experimented with structured learning.
Out of CRF using L-BFGS updates, CRF with passive-aggressive updates to combat Twitter noise~\cite{derczynski2014passive}, and structured perceptron (also useful on Twitter noise~\cite{johannsen2014more}), CRF L-BFGS provided the best performance on our dataset for the ten-types task.

\subsection{Training Data}
In our final system, we included the dev\_2015 data, to combat drift present in the corpus.
We anticipated that the test set would be from 2015.
The original dataset was harvested in 2010, long enough ago to be demonstrably disadvantaged when compared with modern data~\cite{fromreide2014crowdsourcing}, and so it was critical to include something more.
The compensate for the size imbalance -- the dev\_2015 data is 0.175 the size of the 2010 data -- we weighted down the older dataset to by 0.7, as suggested by~\cite{cherry2015unreasonable}, implemented by uniformly scaling individual feature values on older instances.
This successfully reduced the negative impact of the inevitable drift.

\begin{table}
\footnotesize
\centering
\begin{tabular}{lccc}
\hline
{\bf Entity type} & {\bf P} & {\bf R} & {\bf F1} \\
\hline
          company  & 28.07  & 41.03 &  33.33  \\
         facility  & 25.00  & 23.68 &  24.32  \\
          geo-loc  & 53.91  & 53.45 &  53.68  \\
            movie  & 20.00  &  6.67 &  10.00  \\
      musicartist  & 14.29  &  2.44 &   4.17  \\
            other  & 45.78  & 28.79 &  35.35  \\
           person  & 54.63  & 65.50 &  59.57  \\
          product  & 27.78  & 13.51 &  18.18  \\
       sportsteam  & 42.86  & 25.71 &  32.14  \\
           tvshow  &  0.00  &  0.00 &   0.00  \\
\hline
Overall  & 45.72 &  39.64 &  42.46 \\
\hline
No types & 63.81 &  56.28 &  59.81 \\
\hline
\end{tabular}
\caption{Results of the USFD W-NUT 2015 system.}
\label{tab:results}
\end{table}

\section{Performance}

Our results are given in Table~\ref{tab:results}. As can be seen, the best results were achieved for the person and geo-loc entity types. It is also worth noting that performance on the notypes task is significantly better across all metrics, which indicates that the system is capable of identifying entities correctly, but encounters issues with their type classification. 

We found that the biggest contributions to our system's performance were the Freebase gazetteer features, and using Brown clusters with high values of $m$ (the number of classes) and large amounts of recent input data.
This led our computational efforts in the last week to be based around running the biggest Brown clustering task that we could in time.

We also noted during testing that, while passive-aggressive CRF updates helped with single-type entity recognition in tweets~\cite{derczynski2014passive}, it was detrimental to an all-types system.
It was also not helpful for the no-types task, where L-BFGS updates again gave better performance.
This is rational: the all-types and multiple-types tasks are effectively similar when contrasted with the single-types task, in that they require the recognition of many different kinds of named entity.

Finally, we found that other gazetteer types were not helpful to performance; taking for example all of the ANNIE gazetteers, gazetteers from IMDb dumps, entity names extracted from other Twitter NER corpora, or entities generated through LLDA~\cite{Ritter2011} all decreased performance.
We suspect this is due to their swamping already-small input dataset with too great a profusion of information, c.f.~\newcite{smith-osborne:2006:CoNLL-X}.

In addition, we tried generating semi-supervised data using vote-constrained bootstrapping, but this was not helpful either -- presumably due to the initially low performance of machine-learning based tools on Twitter NER making it hard to develop semi-supervised bootstrapped training data, no matter how stringent the filtering of autogenerated examples.

For the final run, we were faced with a decision about fitting.
We could either choose a configuration that minimised training loss on all the available training data (train + dev + dev\_2015), but risked overfitting to it.
Alternatively, we could choose a configuration that fit less well, in order to avoid overfitting.
In the end, we decided to adopt the above principled approach, assuming that final data would be from 2015, and therefore down-weighting training data from prior years.
We also evaluated the system while including the dev\_2015 data in the training set, to see how well we would match it.

\begin{table}
\centering
\footnotesize
\begin{tabular}{lcr}
\hline
{\bf Features}  & {\bf Label} {\bf Weight} \\
\hline
pref=@ & O& 3.368445\\ 
pref=htt & O& 2.049354\\ 
pref=\# & O& 1.979034\\ 
shapeshort-x.x & O& 1.688033\\ 
shapeshort-. & O& 1.552530\\ 
w[-1]=at & B& 1.519609\\ 
p14x11110011111011 & O& 1.326481\\ 
w[-1]=at & O& -1.285570\\ 
w[-1]=of & O& -1.244912\\ 
length-2 & O& 1.196777\\ 
length-3 & O& 1.177138\\ 
shapeshort-x. & O& 1.172663\\ 
in\_gaz=Freebase\_\\
videogameplatform & O& -1.152093\\ 
w[-1]=and & O& -1.143885\\ 
length-1 & B& -1.132128\\ 
shapeshort-Xx & O& -1.128341\\ 
w[-1]=with & O& -1.093224\\ 
w[1]=tonight & O& -1.077982\\ 
shapeshort-0 & B& -1.051406\\ 
\hline
\end{tabular}
\caption{Largest weighted features in notypes model}
\label{tab:strong-feats}
\end{table}

\section{Analysis}

\subsection{Features}
In terms of features, we looked at the strongest-weighted observations in the notypes model, to see what the general indicators are of named entities in tweets.
The largest of these are shown in Table~\ref{tab:strong-feats}.
Of note is that features indicating URLs, hashtags and usernames indicate against an entity; lowercase words including punctuation, or comprising only punctuation, are not entities; being proceeded by {\em at} indicates being in an entity (+ve B weight and -ve O weight); being preceded by {\em of}, {\em and} or {\em with} suggests an entity; short words and hashtag-shaped words are not entities; being followed by {\em tonight} suggests being inside an entity; numbers rarely start entities; and being matched by an entry in the video games gazetteer suggests being an entity.

One cluster prefix was indicative of being outside an entity.
This cluster prefix contained four subclusters, each dominated by lot of frequently-occurring dates (e.g. {\em September} with 12368 mentions in the source data) and less-frequent date spellings like {\em Wedneaday} or rarer occasions {\em Pentecost}, but also a lot of less-frequent noise entries, some of which were potentially named entities (e.g. {\em \#ITV3}, {\em Buggati}, {\em  Katja}).
The noise present suggests that, while the clustering is working well, there are not enough clusters; for 250M tweets, we should use $m>2000$~\cite{derczynski2015brown}.

\begin{table}
\centering
\footnotesize
\begin{tabular}{lcrl}
\hline
{\bf Features}  & {\bf Label} {\bf Weight} & {\bf Terms} \\
\hline
 prev\_p3x011 & B-geo-loc & -0.571505\\ 
 p14x11110011111001 & B-other & -0.585369\\ 
 prev\_p6x111100 & B-company & -0.604976\\ 
 p12x111100111110 & B-geo-loc & -0.620909\\ 
 prev\_p4x0100 & B-person & -0.655420\\ 
 prev\_p18x0000111110 & B-facility & 0.699101\\ 
 prev\_p20x0000111110 & B-facility & 0.699101\\ 
 prev\_p3x010 & B-sportsteam & 0.709865\\ 
 p10x0110011010 & B-tvshow & 0.714127\\ 
 p3x011 & B-person & -0.717037\\ 
 p14x11110011111001 & B-product & 0.747492\\ 
 prev\_p8x11110110 & B-other & 0.774895\\ 
 p14x11110011111100 & B-geo-loc & 0.804635\\ 
 prev\_p3x010 & B-person & -0.894333\\ 
 p12x111100111111 & B-geo-loc & 0.895203\\ 
 p14x11110011110110 & B-person & 0.950866\\ 
 p14x11110011111000 & B-company & 1.044984\\ 
\hline
\end{tabular}
\caption{Largest-weighted Brown cluster features in 10-types task}
\label{tab:strong-brown-feats}
\end{table}

\label{sec:brown}
We also looked at the Brown clusters most indicative of entity starts in the typed task, to get an idea of how these clusters helped.
Results are shown in Table~\ref{tab:strong-brown-feats}.
Without going into too much detail -- the cluster paths are distributed with this work, and on the web,\footnote{See {\small http://derczynski.com/sheffield/resources/gha.250M-c2000.tar}} for further examination -- some top-level observations can be made.
Firstly, the preceding word is often influential; note the large number of {\em prev\_\*} features.
Secondly, the clusters prefixed 111100- contained words often used as the first term in many kinds of entity, suggesting distributional similarities in the first words of named entities. 
As Brown clustering is based on bigram distributionality, this finding aligns with the existence of highly-weighted common preceding tokens seen in the model weights for the notypes task.
Thirdly, Brown clusters are more useful for some entity types than others; there are more features for person, company and geo-loc types than others.

Note the large-weighted shallow-depth features for entities.
One is for the terms found before a sportsteam entity (but not a person, note the -ve weight): {\em prev\_p3x010}.
This cluster subtree contains many adjectives, possessive pronouns and determiners ({\em the}, {\em ur}, {\em dis}, {\em each}, {\em mah}, {\em his} etc.).
The terms helpful when not preceding geo-locs were close to this subtree, differing only in its least-significant bit: {\em prev\_p3x011}.
This other large-weighted shallow-depth feature was also useful for avoiding first terms of person entities.
Its cluster subtree contains common nouns and qualifiers ({\em one}, {\em people}, {\em good}, {\em shit}, {\em day}, {\em great}, {\em little}), though it is not immediately clear how these terms were helpful; perhaps the prominence of this subtree feature is due to its frequency alone, and better regularisation is needed to handle it.

\subsection{Gold standard}
When developing the system, we encountered several problems and inconsistencies in the gold standard. These issues are partly a general problem of developing gold standards, i.e. the more complicated the task is, the more humans tend to disagree on correct answers~\cite{Tissot2015a}. For Twitter NERC with 10 types, some of the tokens are very difficult to label because the context window is very small (140 characters), which then also leads to acronyms being used very frequently to save space, and because world knowledge about sports, music etc. is required.

In particular, the following groups of problems in the gold standard training corpus were identified:

\paragraph{Broad categories:} While some of the NE types are well-defined (e.g. person, geo-loc), other types are very broad and therefore pose a big challenge. This is already evident by the number of gazetteers created per type~(see Table~\ref{tab:ftypes}), i.e. those broad categories consist of many different subtypes. Since the training set is very small, only a handfull of examples are observed for each subtype (e.g. video game), which makes training a classifier for those types very challenging. One of the most challenging types was products, as many different things can be products.

\paragraph{Overlapping types:} Some NEs belong to more than one type, which makes the classification task even more difficult. For example, it is difficult to distinguish between companies and their products with the same name. There are also inconcistent examples of this in the gold standard, e.g. ``{\em I}: O {\em just}: O {\em bought}: O {\em Dior}: B-product {\em mascara}: O''. In this example, ``{\em Dior}'' should be annotated as a company, but ``{\em Dior mascara}'' as a product from that company.

\paragraph{The type \textit{other}:} Since annotation guidelines are not available for the gold standard, we rely entirely on examples in the training set to identify what subtypes belong to the type ``other''. While most examples seem to be public holidays and events, the type also seems to be used for overlapping or otherwise unclear tokens. Examples for this are ``{\em Radio 1}'' (a broadcasting organisation), ``{\em UMASS}'' (a university), ``{\em Edmonton Journal}'' (a broadcasting organisation), ``{\em Dems}'' (democrats, a group of people or a policical party). The type ``other'' is also one for which annotation guidelines differ heavily -- meaning performance does not increase if we try to aggregate the gold standard corpus with over available Twitter NER gold standards.

\paragraph{Inconsistent annotation for hashtags:} Important words in tweets are often preceded by a hashtag to emphasise them, e.g. ``{\em \#JenniferAniston quote of the day}''. Despite the fact that many of the 327 tokens starting with hashtags were named entities, only 5 of them are annotated with NE types (\#Vh1: B-other, \#Astros: B-sportsteam, \#Denver: B-geo-loc, \#Padres: B-sportsteam, \#BB11: B-tvshow). The false negatives belong to different NE types and are mostly easy to spot (e.g. \#BROOKLYN, \#lindsaylohan). A related problem is the annotation of direct mentions of Twitter accounts with @ in sentences, e.g in ``{\em All}: O {\em caught}: O {\em up}: O {\em with}: O {\em @SHO\_weeds}: O {\em !}: O''. In that sentence, ``{\em @SHO\_weeds}'' refers to the Showtime TV series ``{\em Weeds}'' and should be annotated as \textit{tvshow}. However, all tokens starting with @ are annotated as O, so even though this is not neccessarily correct, it is consistent within the gold standard.

\section{Conclusion}
This paper has described the USFD system entered in W-NUT 2015.
It achieves performance through unsupervised feature generation, through Freebase gazetteers, and through weighting input data according to its origin date in order to account for drift.
This lead to state-of-the-art Twitter NER performance.

\section*{Acknowledgments}
This work was partially supported by the European Union under grant agreement No. 611233 {\sc Pheme},\footnote{\texttt{http://www.pheme.eu}} and the UK EPSRC grant No. EP/K017896/1 uComp.\footnote{\texttt{http://www.ucomp.eu}}

\bibliographystyle{aclabbrv}
\bibliography{../../../../../../sale/big,usfd_wnut}

\end{document}